\documentclass[letterpaper, 10pt, conference]{ieeeconf}      

\def\year{2022}\relax

\usepackage{graphicx} 
\usepackage{amsmath} 
\usepackage{amssymb}  
\usepackage{bm}
\usepackage{todonotes}
\usepackage{calrsfs}
\usepackage{xcolor}
\usepackage{multirow}
\usepackage{url}
\usepackage{flushend}

\title{\LARGE \bf
Render-in-the-loop aerial robotics simulator: Case Study on Yield Estimation in Indoor Agriculture
}

\author{Antun Ivanovic, Marsela Polic, Jelena Tabak, Matko Orsag
\thanks{Authors are with 
        Faculty of Electrical Engineering and Computing, 
        University of Zagreb
         {\tt\small [antun.invanovic, marsela.polic, jelena.tabak , matko.orsag] @fer.hr}
        }%
        \thanks{This work has been supported by Croatian Science Foundation under the project Specularia UIP-2017-05-4042}
}

\begin{document}

\maketitle
\thispagestyle{empty}
\pagestyle{empty}

\begin{abstract}
Inspired by recent promising results in sim-to-real transfer in deep learning we built a realistic simulation environment combining a Robot Operating System (ROS)-compatible physics simulator (Gazebo) with Cycles, the realistic production rendering engine from Blender. The proposed simulator pipeline allows us to simulate near-realistic RGB-D images. To showcase the capabilities of the simulator pipeline we propose a case study that focuses on indoor robotic farming. We developed a solution for sweet pepper yield estimation task. Our approach to yield estimation starts with aerial robotics control and trajectory planning, combined with deep learning-based pepper detection, and a clustering approach for counting fruit. The results of this case study show that we can combine real time dynamic simulation with near realistic rendering capabilities to simulate complex robotic systems.
\end{abstract}

\section{Introduction} \label{sec:introduction}
\begin{figure}
	\centering
    \includegraphics[width=0.99\columnwidth]{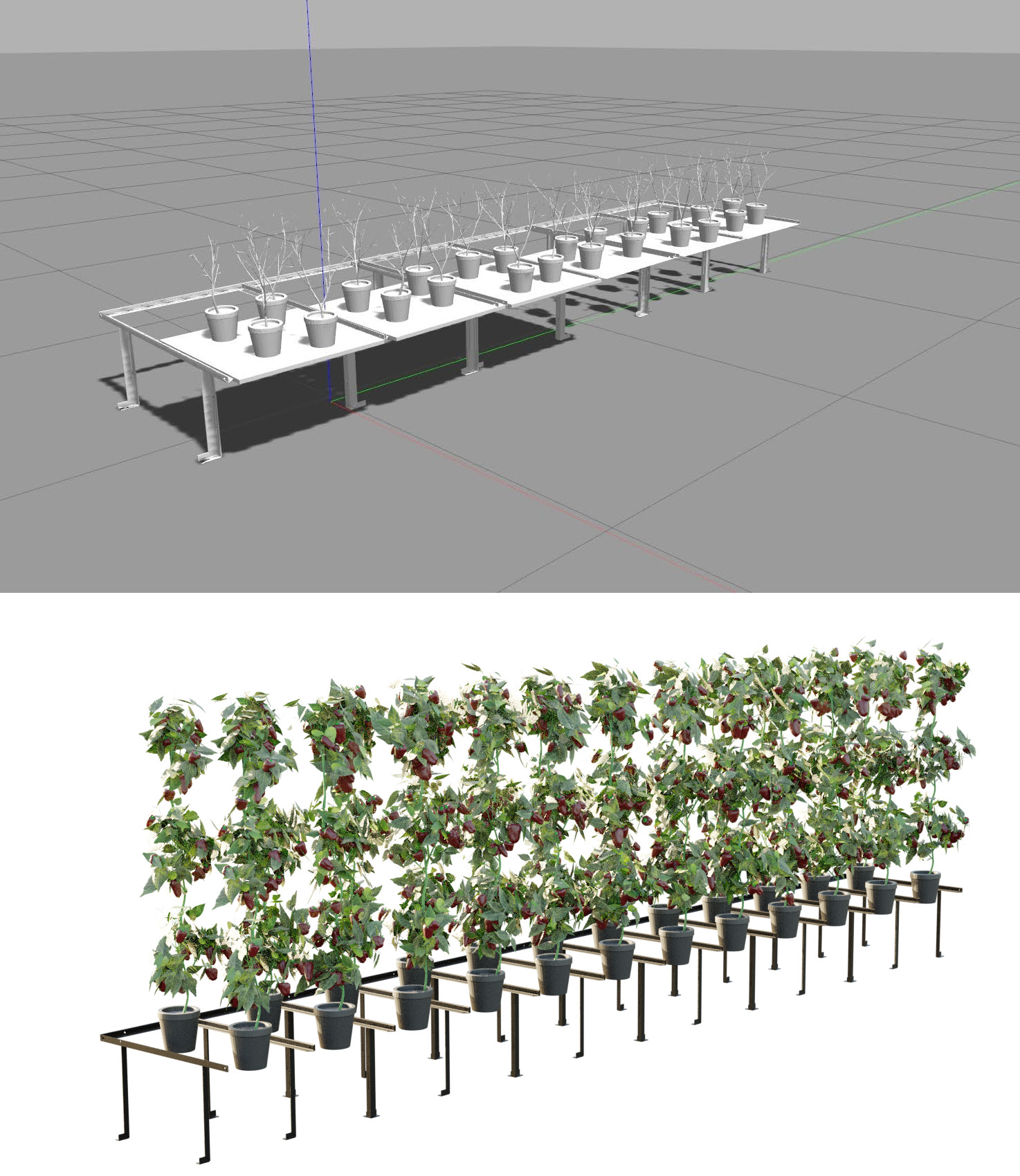}
    \caption{A segment of the greenhouse rendered with the Gazebo simulator, and its respective counterpart rendered in Blender using the Cycles engine. The obvious difference in image quality allows us to apply the same deep learning based pepper detection algorithm we deploy in real world applications. When put together, the two programs enable us to simulate complex mission scenarios, including both low level control and high level vision based algorithms.} 
    \label{fig:blender_vs_gazebo}
 \vspace{-0.55cm}
\end{figure}

Today, as we witness the new dawn of Artificial Intelligence, it is hard to imagine Unmanned Aerial Vehicle (UAV) applications without some form of machine or deep learning. When UAVs are deployed on surveillance, search and rescue, crop monitoring, or infrastructure inspection missions, Convolutional Neural Networks (CNNs) are an ubiquitous part of the solution. In robotics, simulation is the first step towards successfully deployment in real world applications. Real time simulators like Gazebo \cite{Furrer2016}, accurately and efficiently simulate the physics and dynamics of robots, however, the photo-realistic characteristics of images rendered this way leave a great deal to be desired. Arguably, other game based simulators, like Xplane \cite{garcia2009multi}, or  AirSim \cite{pueyo2020cinemairsim}, provide more photo-realistic virtual environments, nevertheless, their photo realism is no match for a dedicated production renderer like Cycles \cite{iraci2013blender}. Combining physics simulation and realistic visual appearance enables us to test the complete aerial robotics solution, starting from the low level control all the way towards the high level mission planning and Artificial Intelligence (AI). The difference between the two render approaches is shown in Fig \ref{fig:blender_vs_gazebo}.

We chose Cycles, from Blender, since it is an open source cross platform solution with an integrated Python API, which makes it straightforward to interface with the ROS environment. Our approach goes hand in hand with recent sim-to-real attempts in deep learning, where photorealistic renderers are used to build large synthetic datasets used to train CNNs. The proposed simulator can be used in different indoor and outdoor scenarios, ranging from infrastructure inspection to crop monitoring. In this work, we focus on the yield estimation for the indoor farming. Indoor farming is becoming an important tool to mitigate the problems farmers face with unpredictable and often extreme weather patterns induced by global warming. Traditional indoor agriculture enables farmers to provide their crops with optimal weather conditions, but outdoor impact, predominantly sunlight, still plays an important role in the plants’ growth. Modern indoor farming completely eliminates the outside weather conditions, emulating optimal weather conditions controlling both the climate and sunlight. 

While greenhouse climate control can be considered as a solved problem, replacing manual labor is an attractive research topic for robotics. From the robotics point of view, indoor farming provides a level of structure in the environment which makes it easier to deploy a fully autonomous operation. The farming industry is extremely labor intensive, and the job requirements often fit the category of dangerous, dull, and dirty, making them ideal for automation. Labor is even more important when considering the organic farming. To reduce the use of pesticides, the organic agriculture requires a lot more manual care, with a comparably smaller agricultural output. The obvious economical consequence of such a production system is a higher cost of organic food. Turning to robotics and automation can help alleviate the costs. 


The heterogeneous team of robots in our project consists of an aerial inspection robot, mobile robots that carry plants around the farm, and a robotic manipulator which treats the plants \cite{polic2021structured}. However, in this paper we focus on the aerial robot trajectory planning in a structured greenhouse environment and fruit counting through the aforementioned Gazebo and Blender pipeline.

\subsubsection*{Contributions}
In this paper we focus on the yield estimation problem and present a pipeline of technologies used to execute such a task and deploy it on the proposed simulation framework. We present an aerial manipulator capable of capturing images close to the plants without harming them. We disseminate the control and trajectory planning framework necessary to deploy such an aerial manipulator. We modified the existing deep learning based sweet pepper detection algorithm to exploit both 2D detection results and 3D scene information acquired with an on-board depth module. We provide the details of the simulation environment used to simulate missions in various conditions. Finally, we provide the results of the simulation analysis using a realistic scenario of the yield estimation for the sweet pepper harvesting.
\subsubsection*{Paper organization}
In the following chapter we describe the kinematics, dynamics and trajectory planning of the aerial manipulator used in the study. Next we present the deep learning based approach in yield estimation. Finally, we show how the proposed simulation pipeline combines the low level and high level control. We conclude the paper with simulation results demonstrating the effectiveness of the proposed render-in-the-loop simulation setup.
\section{Aerial manipulator}

To start, we first present an aerial manipulator equipped with an RGB-D camera at the end-effector shown in Fig \ref{fig:aerial_robot}. Such a vehicle can fly through a structured greenhouse environment while taking images and depth information about the crops, without harming the plants. The RGB-D data gathered during flight is later used for yield estimation. The structured indoor farming environment allows us to assume the position of the UAV is accurately measured. For example, this can be achieved using ultra-wideband technology fused together with simultaneous localisation and mapping \cite{milijas2021comparison}.  



\subsection{Kinematics}
\begin{figure}
	\centering
	\includegraphics[width=\linewidth]{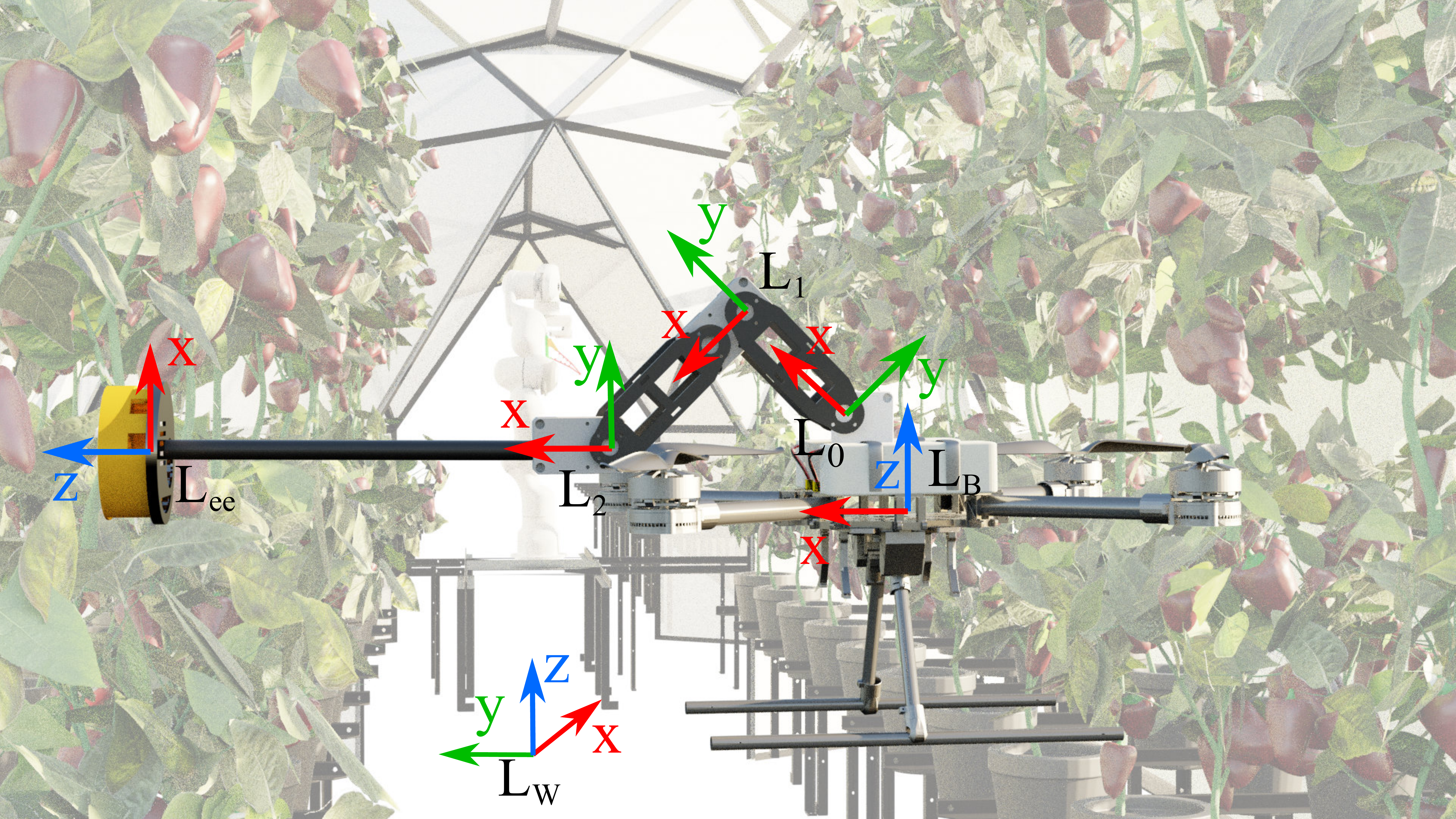}
	\caption{Coordinate systems of the aerial manipulator equipped with an RGB-D camera at the end-effector. The z-axes of the manipulator joints all point into the image.}
	\label{fig:aerial_robot}
\end{figure}

To develop a full kinematic chain of the aerial manipulator, we first define relevant coordinate systems, as shown in Fig. \ref{fig:aerial_robot}. The $L_\textsc{w}$ is the inertial frame, $L_\textsc{b}$ is the body frame placed at the UAV center of mass, $L_0$ is the manipulator base frame, and $L_{ee}$ is the end-effector frame. Let $\mathbf{T}_\textsc{a}^\textsc{b}$ denote a homogeneous coordinate transformation between the coordinate systems $L_\textsc{a}$ and $L_\textsc{b}$. Then we can write a full kinematic chain of the aerial manipulator as:

\begin{equation} \label{eqn:kinematic_chain}
    \mathbf{T}_\textsc{w}^{ee} = \mathbf{T}_\textsc{w}^\textsc{b} \cdot \mathbf{T}_\textsc{b}^\textsc{0} \cdot \mathbf{T}_0^{ee}, 
\end{equation}

\noindent where $\mathbf{T}_\textsc{w}^\textsc{b}$ is the transformation between the world and the UAV body, $\mathbf{T}_\textsc{b}^\textsc{0}$ is a fixed transformation depending on the manipulator mounting point, and $\mathbf{T}_0^{ee}$ can be obtained through the direct kinematics of the manipulator.

In our case, we consider the UAV to have six Degrees of Freedom (DoF), $\mathbf{q}_\textsc{b} = \begin{bmatrix} \mathbf{p}^T & \mathbf{\Theta}^T \end{bmatrix}^T \in \mathbb{R}^6$. The $\mathbf{p} \in \mathbb{R}^3$ denotes the position of the UAV in the world frame, and the $\mathbf{\Theta} = \begin{bmatrix} \phi & \theta & \psi \end{bmatrix}^T \in \mathbb{R}^3$ is the attitude vector containing roll, pitch and yaw angles, respectively. The UAV used in this paper is a multirotor with all rotors placed in one plane. From a control viewpoint, this an underactuated system because only four DoF can be controlled simultaneously. Therefore, we define reduced generalized coordinates of the UAV as $\mathbf{q}_\textsc{b}^* = \begin{bmatrix} \mathbf{p}^T & \psi \end{bmatrix}^T \in \mathbb{R}^4$, where $\psi$ denotes the yaw angle of the UAV. The reduced generalized coordinates will be used later in the planning section \ref{sec:motion_planning}.

We also consider a serial chain manipulator with three DoF attached to the body of the UAV. The generalized coordinates of the manipulator can be written as $\mathbf{q}_\textsc{m} = \begin{bmatrix} q_1 & q_2 & q_3 \end{bmatrix}^T$, where $q_i$ denotes the rotation of each joint. Finally, the full state of the aerial manipulator can be written as $\mathbf{q} = \begin{bmatrix} \mathbf{q}_\textsc{b}^T & \mathbf{q}_\textsc{m}^T \end{bmatrix}^T \in \mathbb{R}^{9}$.

\subsection{Dynamics and control}
In previous section we defined $\mathbf{q}_\textsc{b}$ as generalized coordinates of the UAV. If we take the time derivative we get $\mathbf{\dot{q}}_\textsc{b} = \begin{bmatrix} \mathbf{\dot{p}}^T & \bm{\omega}^T \end{bmatrix}$, where $\mathbf{\dot{p}}$ is the linear velocity and $\bm{\omega} = \begin{bmatrix} \dot{\phi} & \dot{\theta} & \dot{\psi} \end{bmatrix}^T$ is the angular velocity, both expressed in the world frame. Note that the angular velocity is measured in the body frame of the UAV and can be expressed as $\bm{\omega}_\textsc{b} = p\mathbf{x}_\textsc{b} + q\mathbf{y}_\textsc{b} + r\mathbf{z}_\textsc{b}$, where $p$, $q$ and $r$ are angular velocities in the body frame. There exist a simple mapping between the angular velocities in the body frame to the world frame, for more details see \cite{Macdonald2013}. 

In this work we consider using a multirotor vehicle with four rotors. Each rotor produces the thrust force $F_i = k_\textsc{f} \cdot \Omega_i^2$ and the moment $M_i = k_\textsc{m} \cdot \Omega_i^2$ along the $\mathbf{z}_\textsc{b}$ axis, and $\Omega_i$ is the angular velocity of each rotor. As we already mentioned, due to the underactuated nature of the UAV, only four DoF are controllable. Therefore, we can define the UAV control vector $\mathbf{u}_\textsc{uav} = \begin{bmatrix} u_1 & u_2 & u_3 & u_4 \end{bmatrix}^T$, where $u_1$ represents the net thrust and $u_2$, $u_3$ and $u_4$ represent moments around body axes. These control inputs can be written as $\mathbf{u}_\textsc{uav} = \mathbf{K} \cdot \bm{\Omega}_{sq}$, where $\mathbf{K} \in \mathbb{R}^{4 \times 4}$ is the input mapping matrix, and $\bm{\Omega}_{sq} = \begin{bmatrix} \Omega_1^2 & \Omega_2^2 & \Omega_3^2 & \Omega_4^2 \end{bmatrix}^T$ is the vector the squared rotors' angular velocities.

To account for the dynamics of the manipulator, we can write the time derivative of joint angles $\mathbf{\dot{q}}_\textsc{m} = d\mathbf{q}_\textsc{m}/dt$. The control input for the manipulator is a vector of torques for each joint $\mathbf{u}_\textsc{m} = \begin{bmatrix} \tau_1 & \tau_2 & \tau_3 \end{bmatrix}^T$. Combining that with the UAV velocity vector gives the full system velocity $\mathbf{\dot{q}} = \begin{bmatrix} \mathbf{\dot{q}}_\textsc{b}^T & \mathbf{\dot{q}}_\textsc{m}^T \end{bmatrix}^T \in \mathbb{R}^9$. The acceleration vector can be obtained as the time derivative of the velocity $\mathbf{\ddot{q}} = d\mathbf{\dot{q}} / dt$. The full system dynamics can now be written as:

\begin{equation} \label{eqn:dynamics}
\mathbf{M}({\mathbf{q}}){\mathbf{\ddot q}} + {\mathbf{c}}({\mathbf{q}}, {\mathbf{\dot q}}) + {\mathbf{g}}({\mathbf{q}}) = {\mathbf{u}},
\end{equation}

\noindent where $\mathbf{M}({\mathbf{q}}) \in \mathbb{R}^{9 \times 9}$ is the inertia matrix, $\mathbf{c}({\mathbf{q}}, {\mathbf{\dot q}}) \in \mathbb{R}^9$ is the Coriolis vector, $\mathbf{g}({\mathbf{q}}) \in \mathbb{R}^9$ is the gravity vector, and $\mathbf{u} = \begin{bmatrix} \mathbf{u}_\textsc{uav}^T & \mathbf{u}_\textsc{m}^T \end{bmatrix}^T$ is the aerial manipulator control input. Having kinematics and dynamics of the aerial manipulator derived gives us the foundation for the motion planning and simulation discussed in following sections.

\subsection{Aerial manipulator motion planning} \label{sec:motion_planning}
To estimate the number of fruit in the greenhouse, it is necessary to scan and inspect the plants. In this paper, an aerial manipulator with an RGB-D camera mounted on the end-effector is considered for the task. Since the layout of the greenhouse is a-priori known, it is possible to scan a single plant container unit or a whole structured row. The gathered data is then used in the fruit counting algorithm.

\subsubsection{Scan waypoints}
\begin{figure}
	\centering
	\includegraphics[width=\linewidth]{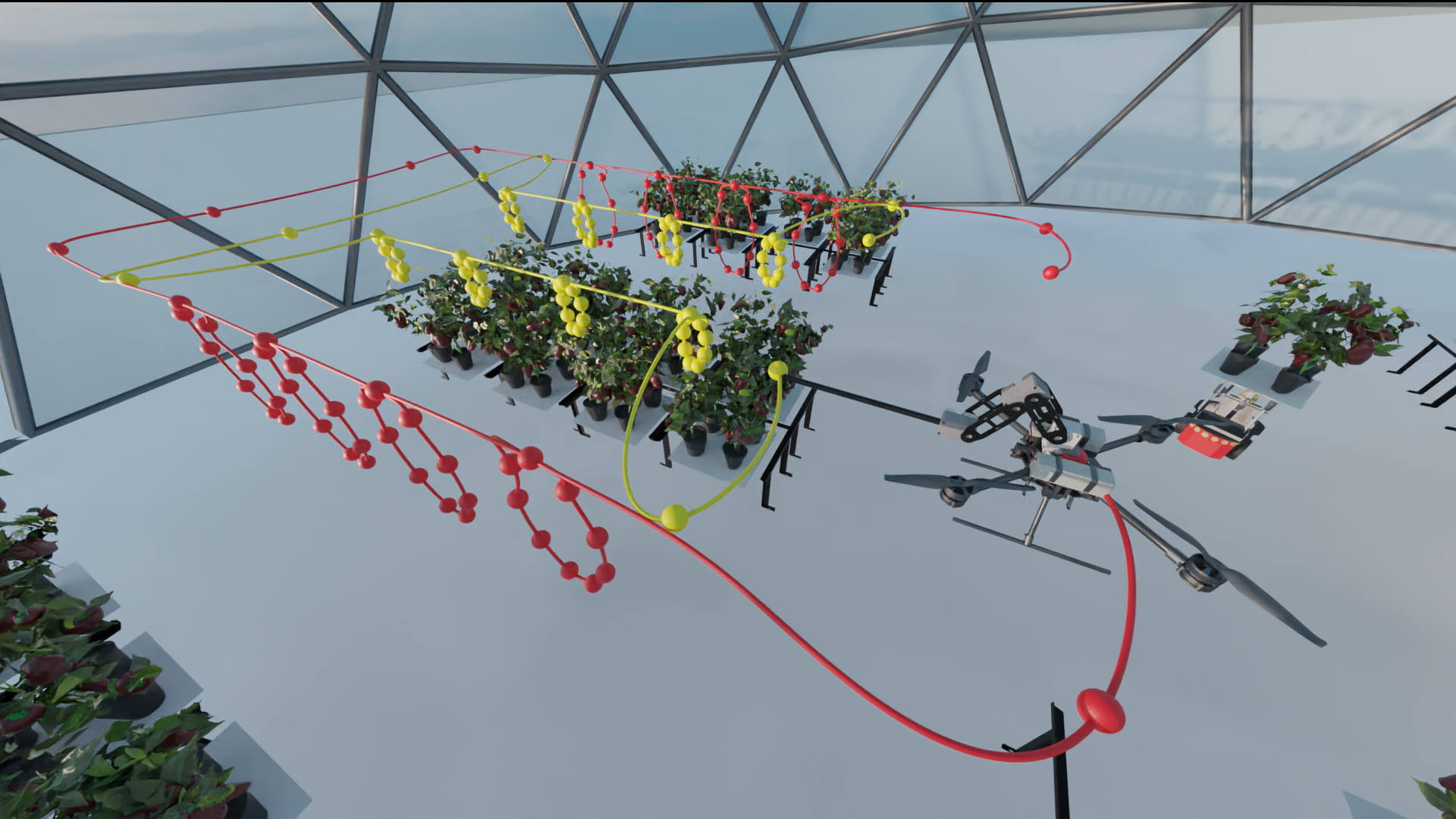}
	\caption{An example of the planned elliptical trajectory used to scan a single row of the structured greenhouse. Red lines and spheres denote the trajectory and waypoints of the UAV body, while the yellow color denotes the end-effector trajectory.}
	\label{fig:waypoints}
\end{figure}

Each plant in a greenhouse row has to be scanned from multiple angles to obtain informative data for fruit counting. An example of planned waypoints and trajectory for a single row is depicted in Fig. \ref{fig:waypoints}. In short, the aerial manipulator moves around the row and scans each plant with an elliptical trajectory.

Precisely scanning each plant requires a carefully planned end-effector motion, which mostly relies on the dimensions of growth containers. The elliptical shape of the waypoints is based on the plant position and dimensions. At each waypoint of the ellipse, the yaw of the UAV directs the end-effector towards the plant's centroid, pointing the end-effector upwards or downwards to get a better overview of the plant.

One row of the greenhouse consists of multiple plants. To perform a scan of a single row, it is necessary to scan each plant from both sides of the row. Therefore, an elliptical set of waypoints is planned for each plant, based on the dimensions and the number of plants in the row. Additional waypoints for navigating around the row are also included to yield a smooth trajectory.

Planned waypoints are considered to form a path of $n$ waypoints:
\begin{equation} \label{eqn:path}
	\mathcal{P} = \left\{\mathbf{q}_i^* \mid \mathbf{q}_i^* \in \mathbb{R}^{7}, i \in (1, 2, \dots, n) \right\},
\end{equation}
\noindent where $\mathbf{q}_i^* = \begin{bmatrix}
(\mathbf{q}_\textsc{b}^*)^T & \mathbf{q}_\textsc{m}^T
\end{bmatrix}^T$ is a single waypoint containing the position and orientation of the UAV, as well as joint positions of the three DoF dimensional manipulator. As discussed in previous sections, both waypoints and path are expressed in terms of reduced generalized coordinates, due to the underactuated nature of the UAV.

\subsubsection{Scan trajectory}
Based on the generated path from equation \eqref{eqn:path}, a trajectory that respects dynamic constraints of the system is planned. Namely, the Time Optimal Path Parameterization by Reachability Analysis (TOPP-RA) algorithm \cite{toppra}, that operates on the numerical integration approach, is employed. As an input, the algorithm requires a set of positions with velocity and acceleration constraints for each DoF. The constraints are dependent on the aerial robot's dynamics and control. The output is a smooth trajectory $\mathcal{T}$:

\begin{equation} \label{eqn:trajectory}
	\mathcal{T} = \left\{ \mathbf{t}(t) \mid \mathbf{t}(t) \in \mathbb{R}^{21}, t \in (0,t_{end}) \right\},
\end{equation}
\noindent where $\mathbf{t}(t) = \begin{bmatrix}
(\mathbf{q}^*)^T & (\mathbf{\dot{q}}^*)^T & (\mathbf{\ddot{q}}^*)^T
\end{bmatrix}^T$ denotes a trajectory point containing position, velocity and acceleration of each DoF, and $t_{end}$ denotes the duration of the trajectory.
\section{Enabling technologies} \label{sec:enabling_technologies}
An important drive to deploy robots in agriculture is just-in-time detection. This is important both for agronomic and economic reasons. Farmers must have an accurate yield estimate of their crops to plan and organize the harvest. In this paragraph we outline the key AI technologies developed to successfully deploy the aerial robot on a sweet pepper yield estimation task.

\subsection{Fruit detection}
As most state of the art solutions, in this work, we rely on a commercial RGB-D camera \cite{fu2020application}. The red peppers are detected in 2D RGB images using a deep learning model. A MobileNet based Single Shot Detector (SSD), pretrained on the COCO dataset \cite{lin2015coco}, is trained for the object detection task in 2D RGB images. For the purposes of network training, a synthetic dataset is procedurally generated in Blender, in order to mitigate the cost of labeling a large training dataset \cite{hinterstoisser2019annotation, khan2019procsy}. Increasingly popular synthetic datasets like \cite{barisic2022sim2air}, have recently found applications in agriculture for various crops and cultures \cite{di2017automatic, olatunji2020reconstruction, zhang2021detecting}, including a synthetic dataset for the \textit{C. annuum} semantic segmentation tasks \cite{barth2018data}. Transfer learning for the network first conducted on the synthetic dataset is followed by additional fine tuning on a small dataset of real, manually labeled images. This encourages us to simulate the entire mission in the proposed simulator setup, before the real world deployment.
 
The 2D detection pipeline produces bounding boxes that denote positions of detected peppers in the image, but the precise spatial information can be obtained combining both detection and depth information of the RGB-D cameras. Various methods have been developed over the recent years for 3D pose estimation, such as surface normal estimation for grasp position optimisation \cite{lehnert2017autonomous}, and peduncle model fitting in harvesting \cite{sa2017peduncle}. In geometric model fitting methods, the detected fruit is modelled with geometric primitives such as cylinders and ellipsoids  \cite{lehnert2016sweet}. In this work, such a high precision is not necessary, hence approximate 3D positions of the detected peppers are obtained applying the 2D detection bounding boxes on the organised pointcloud output from the camera depth channel. From among the filtered points representing a single detected pepper, a centroid point can be chosen as a reasonable approximation of the pepper position. It should be noted that this way, a point on the pepper's surface is chosen, resulting in varying position estimates for the same pepper depending on the camera perspective with respect to the pepper. 

The harvesting mission can be planned based on the ripe fruit count. The aerial manipulator executes a trajectory that enables recording the plants from multiple perspectives. Most of the fruit, unless heavily occluded, is recorded, and detected, from several perspectives during the UAV motion. As stated, these detections do not match perfectly, since the positions of the pepper surface are not at the same global position. A counting method based on unsupervised learning is devised, that provides an estimate of the yield. 

Upon trajectory execution, a set of detections is collected. In these detections, subsets of data are separated for each greenhouse row, using known layout of the greenhouse. Furthermore, with a predefined UAV trajectory, a known camera frame rate, and an estimate of the possible pepper yield, random subsampling is conducted on the detection dataset, filtering out most of false positive detections. For an expected yield of up to 10 peppers uniformly distributed across the plant body, this filtering step retains approximately 5-10 detections of each fruit. The remaining detections are augmented using random noise, to a fixed size set that enables unsupervised learning methods to properly separate the search space. We have found that 500 points are sufficient to properly separate the data, ranging in fruit count to up to 40 peppers per row (10 fruit per plant), with detection variation at centimeter level, and spread over approx. $2 \times 0.5 \times 0.5 \textup{m}$. From among the 500 points, the remaining false positive detections (e.g. pot, table, and other detected outside the plant bounding volumes) are filtered using the known greenhouse layout. In case the greenhouse layout is not a-priori known, outlier removal methods can be used for the false positive filtering \cite{breunig2000lof}. 


\subsection{Ripe fruit count}
On the augmented dataset of detections, the OPTICS algorithm is deployed as a clustering method \cite{ankerst1999optics} that separates the dataset by defining core points. These core points, i.e. cluster centres, are considered as individual peppers. The inputs to the algorithm are maximum distance $\epsilon$, and minimum cluster size $MinPts$. The parameter $\epsilon$ represents the maximum distance of the cluster points to the core point, in order to be considered as a member of the cluster. In our case, this is set to $4$ cm, given the expected size of the pepper fruit. The second parameter, $MinPts$, is a requirement on the cluster size, i.e. on the minimum number of detections to be considered a pepper. Empirical results in manipulator harvesting experiments showed that at least 2 detections from various viewpoints were needed to estimate the pepper position reliably. Due to the dataset augmentation to the fixed size of 500 points, the minimum number of detections is a function of the initial detections dataset size $n_{init}$, i.e. the original requirement of 2 detections increases by the factor of $\left \lceil 500/n_{init} \right \rceil$. The method clusters the points satisfying the provided conditions, and leaves the remaining points undefined.

\begin{figure}[t!]
	\centering
	\includegraphics[width=\linewidth]{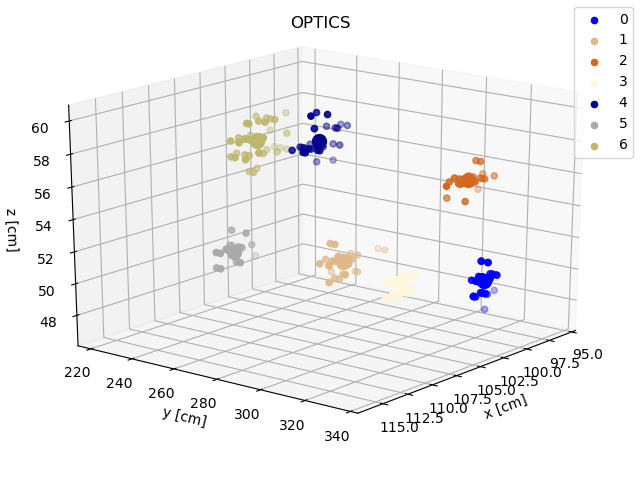}
	\caption{Intermediate results of clustering based pepper fruit counting on a subset of the greenhouse scene. The initial set of detections generated by applying the CNN detector results to the scene point cloud are augmented with a Gaussian random noise until a dataset of fixed size 500 is generated. The OPTICS algorithm separates the augmented search space, identifying 7 clusters (7 peppers). The remaining undefined points are removed for clarity.} \label{fig:clusters}
\end{figure}

Thanks to the organised structure of the greenhouse, the positions of pepper growth containers (pots) are known (or can be known if a similar detection was deployed for pot detection). Although this is not crucial in this paper, it is an important feature for the ultimately envisioned case of autonomous robotic team in a greenhouse environment. This information can be used to select a growth unit to be harvested and the detected pepper positions can be used as a-priori information for harvesting with a manipulator.
\section{Simulation setup} \label{sec:simulation_setup}

\begin{figure}
	\centering
	\includegraphics[width=\linewidth]{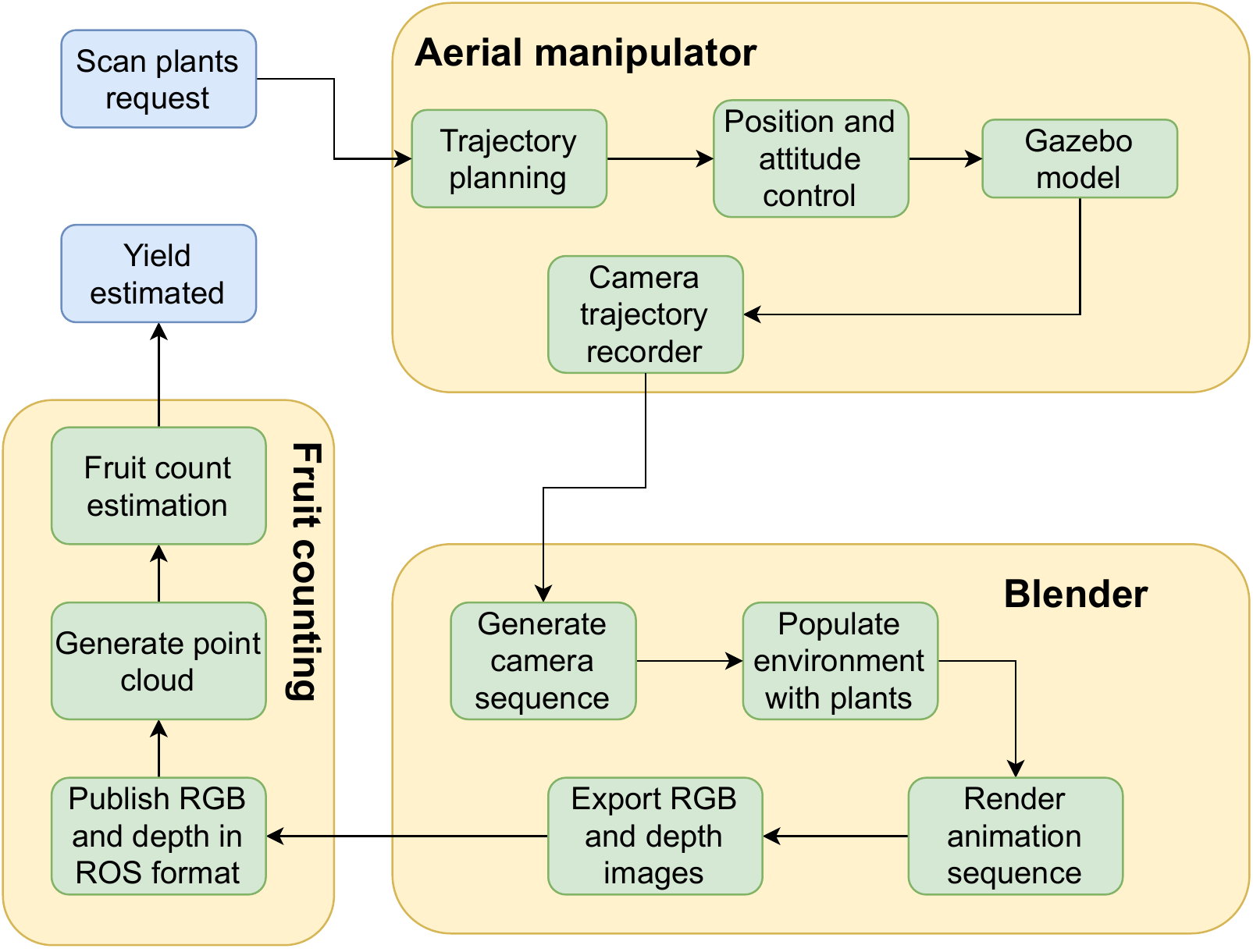}
	\caption{The functional diagram representing the system pipeline. At the  start, a request for scanning the plants is generated, upon which the aerial manipulator executes a scan trajectory in the greenhouse. The gathered camera position and orientation is used through Blender to generate a realistic dataset for fruit counting. The dataset is afterwards processed in ROS to obtain the synchronized point cloud and RGB image for each camera configuration. Using that data, the fruit count is estimated.}
	\label{fig:system_schematic}
\end{figure}

One of the key features which make Gazebo one of the most popular physics simulators in robotics is its simple and straight-forward interface with ROS. However, due to its limitations in photorealistic  visualisation and the complexity of modeling a world containing a large number of detailed models, we propose extending the simulation pipeline with Blender production rendering software, Cycles. The complete pipeline is described with a schematic in Fig. \ref{fig:system_schematic}.

\subsection{ROS/Gazebo}
As depicted with the schematic in Fig. \ref{fig:system_schematic}, for the most part, our solution to the proposed yield estimation is implemented within the ROS environment. This setup enables us to simply switch between the simulation and the real hardware when it is ready to be deployed. The aerial manipulator performs a scanning trajectory, so that the end-effector tip position is used to track the camera motion. The tracked pose of the camera is later imported in Blender, and a photorealistic image of the environment is rendered using Cycles on the exact pose the camera had in the simulation. Based on the rendered camera image, fruit yield is estimated per plant, and provided to the mission planner. The mission planner generates a schedule for the robot(s) involved in the greenhouse cultivation. The speed of the mission execution is a function of available resources and the setup. Mission planning goes beyond the scope of this paper, but interested reader can find more details in \cite{arbanas2021distributed}.

The rotorcraft UAV is simulated in Gazebo using RotorS simulator \cite{Furrer2016} to actuate the body. The UAV is modeled as a rigid body with four propeller units mounted at each of the UAV's arms. The aforementioned three DoF manipulator is attached to the UAV body. More information about this setup and the corresponding GitHub repository is available at \cite{Arbanas2018}. Low level position control both for the UAV body and the manipulator joints are developed in the form of classic cascade PID controllers. 

A simplified 3D model of the greenhouse is used during the Gazebo simulation. Such a setup allows us to simulate the trajectory in real time, while keeping track of potential collisions. Different plugins enable emulating various sensors needed for localization and control. This also allows one to simulate the sensors with variable degree of noise and accuracy, which is important for troubleshooting the solution and it makes the simulation more realistic. 

While performing the simulation, the end-effector configuration is recorded and saved to a file. This is necessary to provide the Blender environment with positions and corresponding orientations of the RGB-D camera for rendering, which is explained in detail in the following section.

\subsection{Blender}
\begin{figure}
	\centering
	\includegraphics[width=\linewidth]{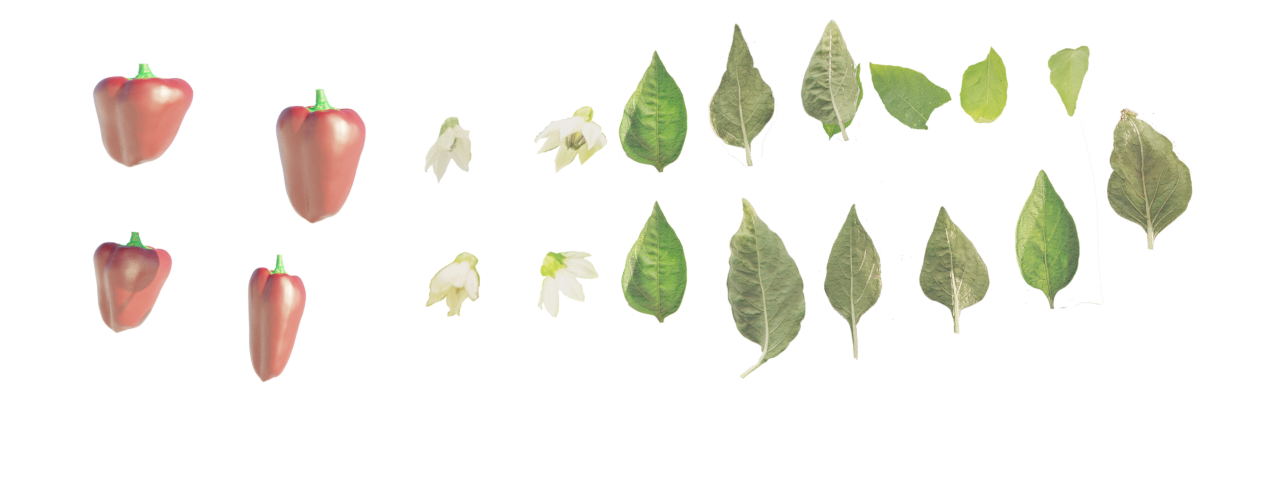}
	\caption{Example models used to build the 3D models of the plants in the simulated greenhouse.}
	\label{fig:3Dassets}
\end{figure}

The model of the greenhouse consists of rows of tables carrying pepper growth containers. Each row consists of 4 tables carrying a single pepper plant. The plants are modeled using the basic building blocks developed for synthetic dataset generation, namely realistic pepper models, planar leaf models with realistic textures, flower models, plant stem, and a plastic pot. Additional advantage Blender has over Gazebo at this point is that the plants are generated procedurally with a random particle generator. This means that in different trials one can vary the pepper, the flower and the leaf number, position, and orientation with respect to the plant stem. More info on the setup we use to train the AI to detect peppers can be found in \cite{polic2021pepper}, where we successfully demonstrated pepper detection method trained using both synthetic and actual pepper images. The 3D assets used to build this synthetic dataset were used to create random 3D models of the greenhouse. Example of these 3D models are shown in Fig \ref{fig:3Dassets}.

Once the trajectory is simulated using the simplified model in Gazebo, we record the position and orientation of the RGB-D camera. Camera poses are imported in Blender, where each sample from the Gazebo simulation is rendered using Cycles. Since we can vary the number of peppers and the shape of the plants, this setup allows us to obtain multiple variable attempts using the same trajectory generated in the Gazebo.

\subsection{Simulated RGB-D camera }
Another important role Blender plays in this paper is the animated camera system, simulating an RGB-D sensor for the Gazebo robot simulation. The simulation pipeline links the motion of the end-effector of the aerial manipulator, with an eye-in-hand RGB-D camera, to camera animation in the Blender environment model. The RGB image of the scene is rendered using built-in Blender renderer, Cycles. The depth image is generated from the depth map calculated within Blender Cycles rendering engine. The depth map is a part of the depth buffer, also known as a z-buffer, usually used in computer graphics to represent depth information of objects from a particular perspective \cite{theoharis2001magic}. Cycles is a ray tracing computer graphics algorithm that uses depth buffers to render a scene ensuring that the correct polygons are shown from the provided camera perspective. 

Modifying the algorithm in \cite{Javonne2019}, we were able to render depth and RGB images from camera viewpoints and corresponding camera positions. This data is afterwards processed offline in ROS, transforming the recorded depth image into an organized point cloud \cite{Aivanovic2021}. This framework loads RGB and depth image and publishes them sequentially, together with the intrinsic camera parameters, camera position and orientation. We rely on the standard \textit{depth\_image\_proc} ROS package that generates point cloud from the depth image, and publishes it accordingly. The point cloud is published along with the corresponding RGB image and global camera transformation. This format corresponds to the output of a real RGB-D camera, and can be used by the detection and yield estimation software solution. Extrinsic parameters of the camera are required so that the pepper detection and clustering algorithm can estimate the yield of the simulated greenhouse. The detection and counting is implemented in Python, relying on the TensorFlow and Pythons Sklearn module.

\section{Simulation results and conclusion} \label{sec:experimental_results}

\begin{figure}
    \centering
    \includegraphics[width=\linewidth]{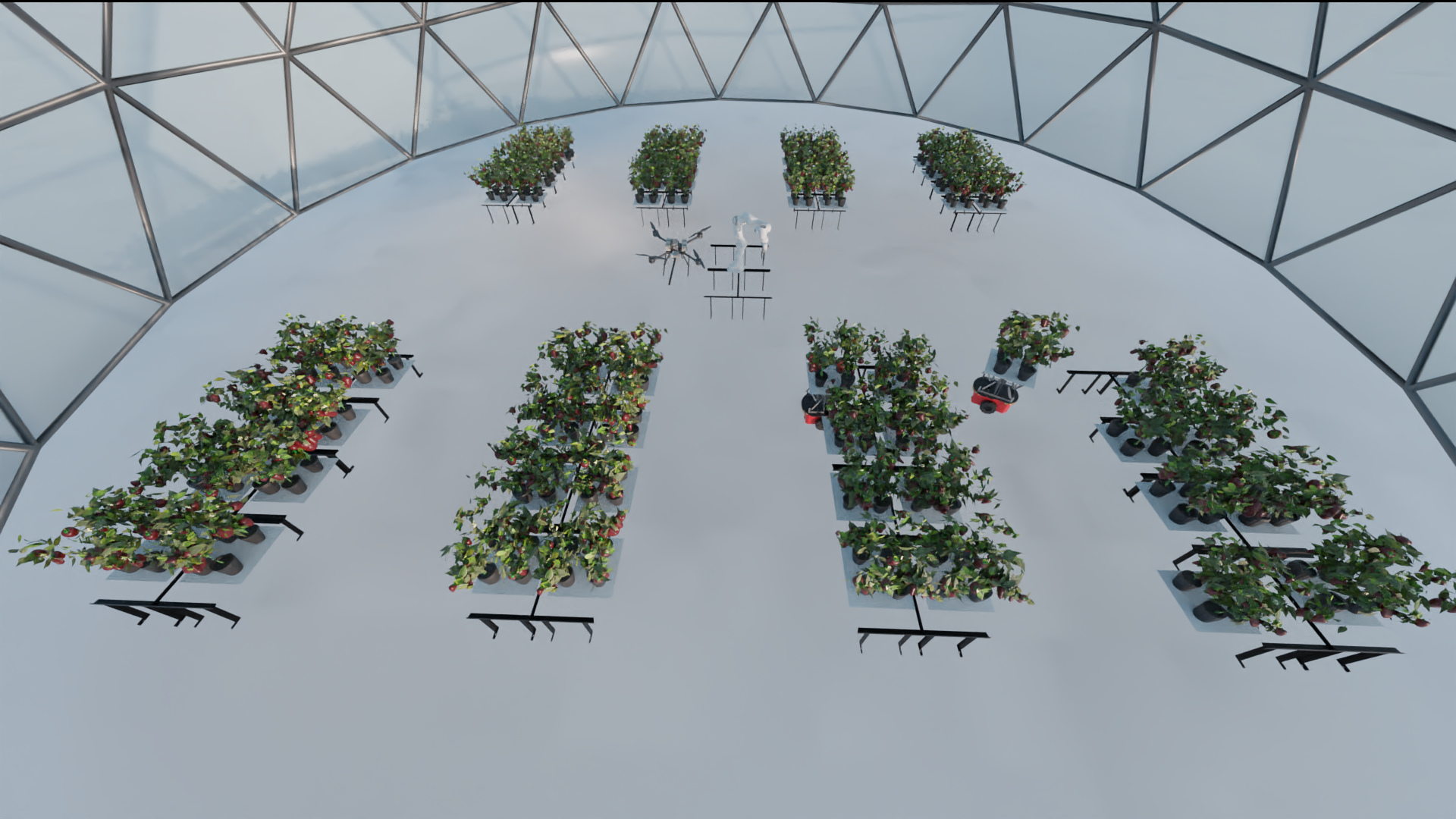}
    \caption{The layout of the proposed greenhouse structure consisting of eight tables, each with four pepper plant containers. The stationary manipulator workstation is located in the center of the structure.}
    \label{fig:greenhouse-layout}
\end{figure}

The greenhouse layout we used for simulation is shown in Fig. \ref{fig:greenhouse-layout}. The structure consists of eight tables, each with four pepper plant containers. The stationary manipulator workstation is located in the center of the structure. For this layout, we ran several simulations with a random number of peppers per plant. For every run the plants were generated randomly. We tested the entire proposed pipeline generating the UAV trajectory for greenhouse inspection, counting the fruits from the video rendered in Blender, and planning the mission based on the inputs from the pepper counting.

To test the planner with a random number of fruits per pepper plant, we selected $20$ of the $40$ available plants that bore fruit. For these plants, a random number of peppers between $1$ and $5$ was generated from a uniform distribution for the left and right sides of the plant. The simulation setups for this use case in terms of number of peppers ranged from 3 to 6 peepers per plant. For every scenario the yield estimation was correct with up to 10\% margin of error.

The results show that \emph{Render-in-the-loop} setup can be used to simulate both the dynamics of the system and vision algorithms for the proposed yield estimation scenario. The complexity of the proposed scenario allows us to assume this simulator pipeline can be used in different scenarios in various indoor and outdoor environments. Future work includes simulating various infrastructure inspection missions as well as outdoor agriculture applications.



\bibliographystyle{ieeetr}
\bibliography{bib.bib}

\end{document}